\documentclass{article} 
\usepackage{iclr2027_conference,times}

\usepackage{amsmath,amsfonts,bm}

\def\eqref#1{equation~\ref{#1}}

\def\1{\bm{1}}

\DeclareMathAlphabet{\mathsfit}{\encodingdefault}{\sfdefault}{m}{sl}
\SetMathAlphabet{\mathsfit}{bold}{\encodingdefault}{\sfdefault}{bx}{n}

\usepackage{url}
\usepackage{booktabs}
\usepackage{graphicx}
\usepackage{amssymb}
\usepackage{multirow}
\usepackage{colortbl}  
\usepackage{xcolor}
\usepackage[colorlinks=true,allcolors=blue]{hyperref}

\title{Latent Bridge: From Video Latents to 3D Geometry via Feature Alignment}

\author{%
\parbox[t]{\textwidth}{%
\raggedright
\textbf{Shuai Zhang\textsuperscript{1}, Zhanpeng Ouyang\textsuperscript{2,$\ddagger$}, Zhengqing Chen\textsuperscript{2},}
\textbf{Ruohong Mei\textsuperscript{2}, Shaoyu Chen\textsuperscript{2}, Weiqiang Ren\textsuperscript{2}, Hui Sun\textsuperscript{2},}
\textbf{Qian Zhang\textsuperscript{2}, Wenyu Liu\textsuperscript{1}, Xinggang Wang\textsuperscript{1,$\dagger$}}\\[5pt]
{\normalfont\textsuperscript{1}Huazhong University of Science and Technology} 
{\normalfont\textsuperscript{2}Horizon Robotics}\\
{\normalfont\texttt{\{shuaizhang,xgwang\}@hust.edu.cn}}\\[4pt]
{\normalfont
\textsuperscript{$\ddagger$}Project leader
\quad
\textsuperscript{$\dagger$}Corresponding author}
}%
}

\iclrfinalcopy 
\begin{document}

\begingroup
\setlength{\tabcolsep}{0pt}
\maketitle
\endgroup

\begin{abstract}
Video generation and feed-forward 3D reconstruction typically use separate representations, requiring RGB decoding and visual re-encoding to recover geometry from generated latents. We propose Latent Bridge, which connects video generative latent spaces with geometric feature spaces to directly predict 3D geometry from video latents. We develop representation-specific bridges for Wan2.1 and V-RAE, enabling feed-forward prediction of cameras, depth, and 3D point clouds without the intermediate RGB pathway. Each bridge is trained on RGB videos through feature distillation, while the video encoder and geometry model remain frozen. Training requires neither ground-truth geometry nor execution of the downstream geometric aggregator or prediction heads. At inference, the same video latents support both RGB decoding and geometry prediction. The same trained bridge can be reused for image-to-video and text-to-video generation within the same video latent space, without bridge retraining. Experiments on Co3Dv2 and WildRGB-D demonstrate competitive geometric reconstruction quality, efficient end-to-end generation
and reconstruction, and faster latent-to-geometry decoding,
highlighting latent alignment as a practical and reusable
interface between video generation and 3D reconstruction. Our code is available at: https://github.com/hustvl/Latent-Bridge.

\end{abstract}

\section{Introduction}

Recent advances in video generation and feed-forward 3D reconstruction
have enabled compelling visual synthesis and efficient geometric inference.
Video generation models, such as CogVideoX~\citep{yang2025cogvideox}
and Wan2.1~\citep{wan2025wan}, combine spatiotemporally compressed
latent representations with diffusion transformers to synthesize
high-quality, temporally coherent videos.
Meanwhile, feed-forward reconstruction models, such as
VGGT~\citep{wang2025vggt}, jointly predict camera parameters,
depth maps, and 3D point maps from multi-view images in a single forward pass.
Recent work has begun to explore the joint execution of these two tasks: generating videos while simultaneously reconstructing the generated content.

Despite their complementary goals, video generation and feed-forward
3D reconstruction typically rely on different representations.
Video generators synthesize content in spatiotemporally compressed
latent spaces, whereas reconstruction models often aggregate
features from visual foundation models such as
DINOv2~\citep{oquab2024dinov} to infer scene geometry.
This representational gap means that combining the two models
typically involves decoding generated latents into RGB frames
and re-encoding them into geometric features, introducing
additional computation. This raises the question: \textit{Can scene geometry be generated directly without RGB decoding?}

\begin{figure*}[t]
	\centering
	\includegraphics[width=1\textwidth]{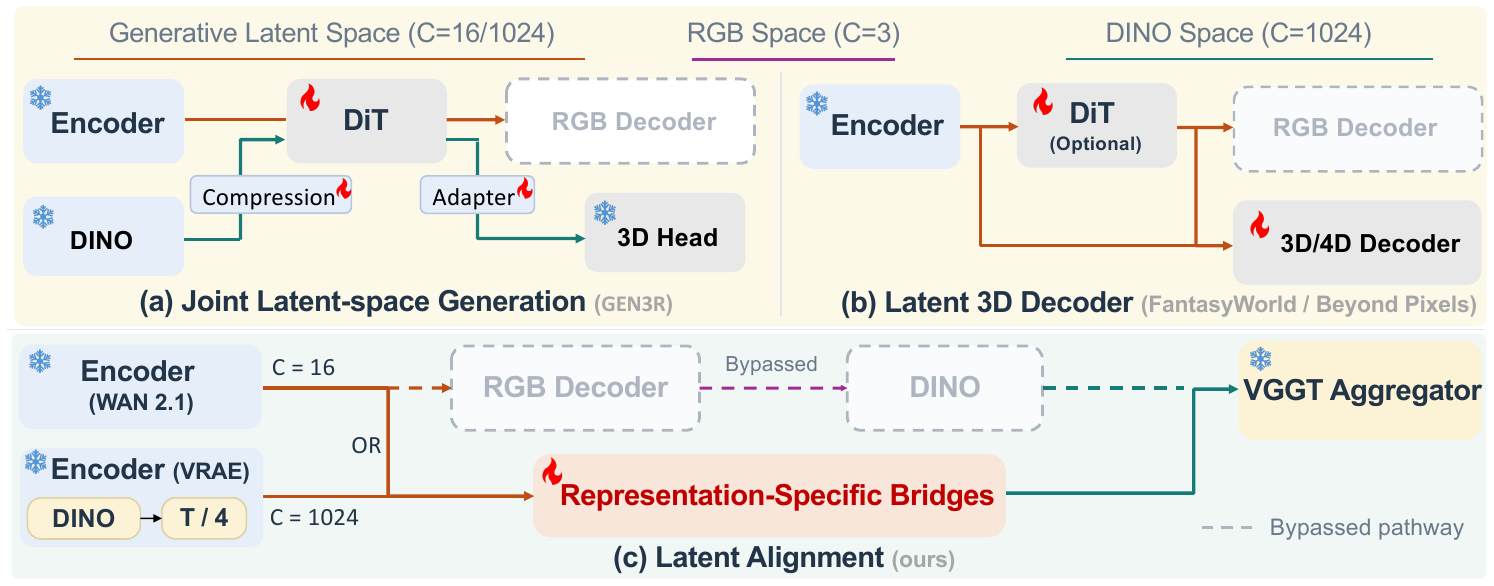}
    \vspace{-10pt}
	\caption{Bypass the video decoder and directly align latent spaces to generate 3D outputs.}
	\label{fig:fig1}
    \vspace{-10pt}
\end{figure*}

Recent work has begun to integrate video generation with explicit geometry prediction. Gen3R~\citep{huang2026gen3r} compresses intermediate VGGT features into geometric latents, aligns their distribution with video appearance latents, and fine-tunes a video diffusion model to jointly generate both. As shown in Fig.~\ref{fig:fig1}(a), joint generation in a shared latent space requires additional training and can reduce efficiency due to the increased number of tokens.
FantasyWorld~\citep{dai2026fantasyworld} introduces a geometric branch into video generation. VGGRPO~\citep{an2026vggrpo} uses latent-space geometry predictions to construct rewards for video model post-training, while Beyond Pixels~\citep{liu2026beyond} combines latent alignment, LoRA adaptation, and geometry-supervised fine-tuning of prediction heads for direct latent-to-4D prediction. As shown in Fig.~\ref{fig:fig1}(b), these methods all aim to train a 3D/4D decoder module, requiring substantial data and retraining of the decoder, while also raising concerns about generalization.


Unlike the aforementioned methods, we argue that the key lies in aligning the generative and reconstruction latent spaces. We propose Latent Bridge, which directly maps video latents into the representation space of a frozen feed-forward geometry model (Fig.~\ref{fig:fig1}(c)). The frozen downstream VGGT aggregator and depth head can then be used to reconstruct the scene. This provides an elegant solution that eliminates the need for joint feature training or training downstream decoder heads. It also generalizes well, enabling the reuse of existing generative models and downstream tasks built on the same representation. Specifically, we adapt our framework to two representative video encoders, the Wan2.1 VAE and V-RAE~\citep{guo2026v}, by designing a dedicated Bridge network for each. This enables models built upon either encoder to directly generate scene geometry.

Our contributions can be summarized as follows:
\begin{itemize}
    \item We propose Latent Bridge, which aligns the generative latent space with the representation space of a reconstruction model, enabling geometry to be recovered directly from generated latents. The Bridge module is trained through feature distillation and requires only video data for training. This substantially reduces the cost of recovering geometry from the latent spaces of generative models.
    \item We adapt our framework to two video encoders, Wan2.1 and V-RAE, by designing a dedicated Latent Bridge network for each. Both variants achieve strong performance, demonstrating that the Latent Bridge framework can effectively accommodate different generative latent spaces.
    \item Experimental results demonstrate that the proposed Latent Bridge enables high-quality and efficient reconstruction. It also exhibits strong generalization: within the same latent space, it can support generative models designed for different generation tasks, as well as other downstream tasks, without retraining.
    
\end{itemize}

\section{Related Work}

\paragraph{Generative latent space.}
Modern video generation models commonly extend latent diffusion to the
spatiotemporal domain, where a video VAE compresses RGB frames into compact
latent representations and a diffusion transformer models their temporal
dynamics. Stable Video Diffusion~\citep{blattmann2023stable} establishes a
scalable latent video diffusion framework, while
HunyuanVideo~\citep{kong2024hunyuanvideo} demonstrates the effectiveness of
large diffusion transformers and progressive training for high-fidelity video
generation. Wan2.1~\citep{wan2025wan} follows this paradigm and combines a
causal spatiotemporal VAE with a diffusion transformer to support both
text-to-video and image-to-video generation. 
REPA~\citep{REPA} and VA-VAE~\citep{yao2025reconstruction}
align diffusion features and VAE latents, respectively, with pretrained
visual representations to improve generative modeling.
RAE~\citep{zheng2026diffusion} directly adopts frozen visual encoders,
while V-RAE~\citep{guo2026v} extends this paradigm to video through
temporal pooling.

\paragraph{Feed-forward reconstruction.}
DUSt3R~\citep{wang2024dust3r} formulates 3D reconstruction as pointmap
regression, jointly predicting dense 3D points for an image pair in a
shared coordinate frame without requiring known camera parameters.
Moving beyond pairwise prediction,
VGGT~\citep{wang2025vggt}, built upon a DINOv2-pretrained visual
encoder~\citep{oquab2024dinov}, jointly predicts camera parameters,
depth maps, 3D point maps, and point tracks from multiple RGB images
in a single forward pass, without per-scene optimization. VGGT-$\Omega$~\citep{wang2026vggtomega}
further initializes its trainable visual backbone from
DINOv3~\citep{dinov3} and introduces scene registers, register
attention, and a shared dense prediction head. Together with large-scale
self-supervised video training, these designs improve the scalability and
efficiency of feed-forward geometry reconstruction for both static and
dynamic scenes.

\paragraph{Joint generative-reconstructive method}
Recent work has explored video and 3D co-generation through complementary approaches. FantasyWorld~\citep{dai2026fantasyworld} couples video and geometry latents through a trainable geometry branch on a frozen Wan2.1 backbone, while LYRA~\citep{bahmani2026lyra} introduces a self-distilled 3D Gaussian decoder.  Gen3R~\citep{huang2026gen3r} jointly generates video and geometry in a unified latent space, whereas PixWorld~\citep{gao2026pixworld} operates in pixel space to avoid VAE-induced detail loss. VGGT-World~\citep{sun2026vggt} instead forecasts geometry-centric VGGT tokens and decodes future geometry using frozen heads, without synthesizing RGB frames. VGGRPO~\citep{an2026vggrpo} uses latent-space geometry predictions to construct rewards for video DiT post-training, while Beyond Pixels~\citep{liu2026beyond} directly decodes geometry from video latents. VIST3A~\citep{go2026texttod} connects video latents to a selected intermediate reconstruction layer and jointly trains the stitching layer and downstream LoRA adapters using teacher-predicted geometry. Pano2World~\citep{li2026pano2world} aligns diffusion hidden features with a pretrained reconstruction model's token space through feature distillation, followed by joint fine-tuning of the adapter and Gaussian decoder under rendering supervision. In contrast, we map video latents to a predefined geometric feature interface through feature alignment, keeping the downstream geometric aggregator and prediction heads entirely frozen.

\section{Preliminaries}
\label{sec:preliminaries}

\subsection{Video Latent Representations}
\label{sec:video_latents}

Video generation models typically use an encoder to compress videos into a latent space. We define an RGB video with \(T\) frames, height \(H\), and width \(W\) as \(\mathbf{X}\in\mathbb{R}^{T\times3\times H\times W}\).
In this paper, we focus on two types of VAEs: the convolutional VAE used in Wan2.1~\citep{wan2025wan} and the representation-based autoencoder V-RAE~\citep{guo2026v}.

\paragraph{Wan2.1 VAE.}
Wan2.1~\citep{wan2025wan} employs a causal spatiotemporal VAE with a spatial compression factor of \(8\) and \(16\) latent channels.
The first frame is compressed only spatially, while subsequent
frames undergo an additional temporal downsampling factor of $4$.
For $T-1$ divisible by $4$, its encoder produces
\begin{equation}
    \mathbf{Z}_{\mathrm{W}}
    = E_{\mathrm{W}}(\mathbf{X})
    \in
    \mathbb{R}^{T_{\mathrm{W}}\times16\times(H/8)\times(W/8)},
    \qquad
    T_{\mathrm{W}}=1+\frac{T-1}{4},
    \label{eq:wan_latent}
\end{equation}
where $E_{\mathrm{W}}$ denotes the encoding operation and
$T_{\mathrm{W}}$ is the latent temporal length.
The corresponding decoder reconstructs RGB frames from
$\mathbf{Z}_{\mathrm{W}}$.

\paragraph{V-RAE.}
V-RAE~\citep{guo2026v} instead constructs its latent space using a frozen visual representation encoder. We consider its DINOv3-based variant, which extracts frame-wise features with $1024$ channels and a spatial compression factor of $16$. Throughout this paper, V-RAE refers specifically to the pretrained weights of this DINOv3-based variant.
A learned temporal attention pooler aggregates adjacent
feature maps while preserving their channel dimension:
\begin{equation}
    \mathbf{Z}_{\mathrm{R}}
    = P_{\mathrm{R}}\!\left(E_{\mathrm{R}}(\mathbf{X})\right)
    \in
    \mathbb{R}^{T_{\mathrm{R}}\times1024\times(H/16)\times(W/16)},
    \label{eq:vrae_latent}
\end{equation}
where $E_{\mathrm{R}}$ is the frozen representation encoder,
$P_{\mathrm{R}}$ is the temporal pooler, and $T_{\mathrm{R}}$
denotes the pooled temporal length.
With four-frame pooling, each latent step summarizes a
four-frame feature window, yielding \(T_{\mathrm{R}}=T/4\).
A spatiotemporal Transformer decoder maps these representations
back to RGB videos.
Unlike the low-channel VAE bottleneck, V-RAE retains
high-dimensional visual features as its generative representation.

\subsection{Feed-Forward Geometry Reconstruction}
\label{sec:geometry_preliminaries}

VGGT~\citep{wang2025vggt} extracts DINO-based image tokens
and aggregates them through alternating frame-wise and
global attention.
Its prediction heads estimate camera parameters, depth maps,
and point maps.
VGGT-$\Omega$~\citep{wang2026vggtomega} introduces register
attention for efficient cross-frame information exchange
and simplifies dense prediction to a single depth head,
alongside a camera head.
For the camera--depth pathway:
\begin{equation}
    \left\{
    \mathbf{D}_t,\mathbf{K}_t,\mathbf{R}_t,\mathbf{t}_t
    \right\}_{t=1}^{T}
    =
    \mathcal{H}_{\mathrm{geo}}
    \!\left(
    \mathcal{A}_{\mathrm{geo}}
    \!\left(E_{\mathrm{img}}(\mathbf{X})\right)
    \right),
    \label{eq:geometry_prediction}
\end{equation}
where $E_{\mathrm{img}}$ is the image feature encoder,
$\mathcal{A}_{\mathrm{geo}}$ is the multi-frame aggregator,
and $\mathcal{H}_{\mathrm{geo}}$ denotes the prediction heads
and camera-parameter conversion.
Here, $\mathbf{D}_t\in\mathbb{R}^{H\times W}$ is the depth map,
$\mathbf{K}_t\in\mathbb{R}^{3\times3}$ is the intrinsic matrix,
and $\mathbf{R}_t\in\mathbb{R}^{3\times3}$ and
$\mathbf{t}_t\in\mathbb{R}^{3}$ define the world-to-camera
rotation and translation. The predicted depth maps and camera parameters can be used to recover
3D point clouds through standard depth unprojection.

\section{Method}

\begin{figure*}[th]
	\centering
	\includegraphics[width=1\textwidth]{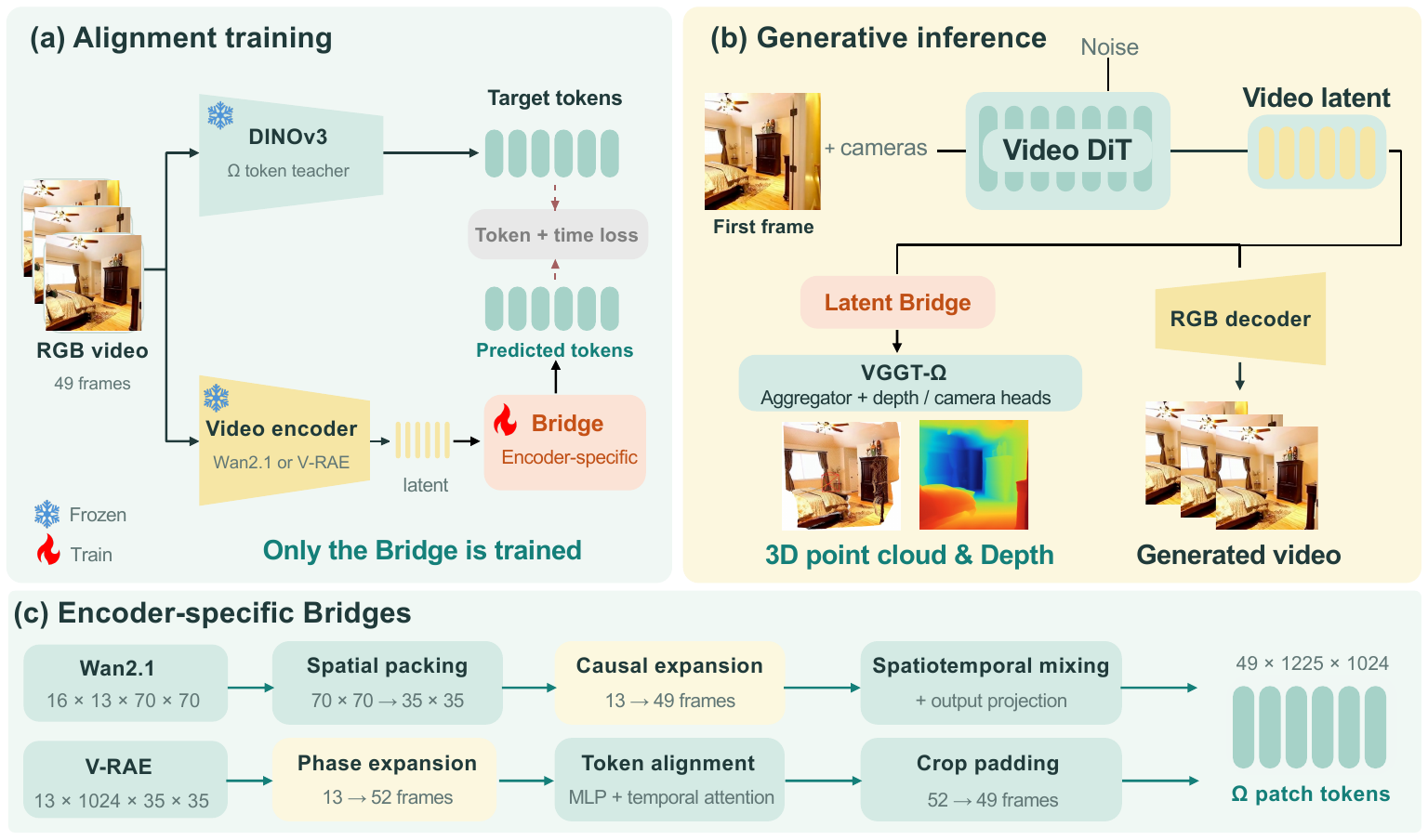}
	\caption{Overview of Latent Bridge, showing (a) feature alignment training, (b) video generation and geometry prediction from a shared latent, and (c) the architectures of bridges.}
	\label{fig:fig2}
    \vspace{-10pt}
\end{figure*}

We propose an alignment framework between generative latents and the representations of a feed-forward geometry model, which we term the \textbf{Latent Bridge}. The key idea is to jointly support video generation and 3D reconstruction by aligning generative and reconstruction representations, without retraining the existing vision autoencoder, generative model, or 3D reconstruction heads. For the Wan2.1 VAE and V-RAE, we design distinct Latent Bridge networks and develop a corresponding training procedure, as illustrated in Fig.~\ref{fig:fig2}. At inference, the bridge enables the frozen geometry model to predict 3D geometry directly from generated video latents.

\subsection{Generative–reconstruction latent-space alignment.}

Given an RGB video
\(\mathbf{X}\in\mathbb{R}^{T\times3\times H\times W}\), we consider a frozen
video encoder \(E_e\), where \(e\in\{\mathrm{V},\mathrm{A}\}\) identifies the
V-RAE or Wan representation. The encoder produces a compressed video latent
\(\mathbf{Z}^{e}=E_e(\mathbf{X})\). 
An encoder-specific bridge \(B_{\theta_e}\) maps this latent to frame-level tokens that match the outputs of the frozen VGGT-$\Omega$ patch encoder \(\Phi_{\Omega}\):

\begin{equation}
\begin{aligned}
\widehat{\mathbf{Y}}^{e}
    &=B_{\theta_e}\left(E_e(\mathbf{X})\right)
      \in\mathbb{R}^{T\times N\times C},\\
\mathbf{Y}
    &=\Phi_{\Omega}(\mathbf{X})
      \in\mathbb{R}^{T\times N\times C},
    \qquad e\in\{\mathrm{V},\mathrm{A}\},
\end{aligned}
\label{eq:unified_latent_alignment}
\end{equation}
\vspace{-10pt}

where \(\widehat{\mathbf{Y}}^{e}\) denotes the tokens predicted from latent
representation \(e\), and \(\mathbf{Y}\) denotes the target tokens extracted
from RGB frames by the frozen Omega patch encoder \(\Phi_{\Omega}\). Here,
\(T\) is the number of frames, \(N\) is the number of spatial tokens per frame,
and \(C\) is the token dimension. For videos with \(T=49\) frames and spatial
resolution \(H=W=560\), the native Omega representation contains
\(N=35\times35=1225\) tokens per frame with \(C=1024\) channels. Our objective is to align \(\widehat{\mathbf{Y}}^{e}\) with \(\mathbf{Y}\).
\(B_{\theta_V}\)and \(B_{\theta_A}\) do not share parameters because their input representations
have different tensor structures. Nevertheless, they share the same target
token space, alignment objective, and frozen geometry decoder.

\subsection{Encoder-Specific Latent Bridges}

\paragraph{Wan Raw-Latent Bridge.}
The Wan branch operates directly on the raw output of the frozen causal video
VAE. For a 49-frame 560$\times$ 560 video, the Wan encoder produces
\(\mathbf{Z}^{\mathrm{A}}=E_A(\mathbf{X}) \in
\mathbb{R}^{T_z\times H_z\times W_z\times C_z}\), where
\(C_z=16\), \(T_z=13\), and \(H_z=W_z=70\). The target dimension is \(\widehat{\mathbf{Y}}^{A}\in \mathbb{R}^{T\times N \times C}\), where
\(C=1024\), \(T=49\), and \(N=(H/16 \times W/16)=35 \times35\). Therefore, we need to spatially downsample the features, restore the temporal dimension to 49 frames, and map them from a lower-dimensional space to a higher-dimensional one.

As shown in Fig.~2(c), the Wan bridge first applies
$2\times2$ spatial packing via PixelUnshuffle to map the
$70\times70$ latent grid to the $35\times35$ VGGT-$\Omega$
patch grid, followed by a learned channel projection.
Causal phase expansion, implemented by a causal 3D convolution
followed by channel-to-time phase rearrangement, maps
$T_z=13$ latent steps to $T=49$ frame positions according to
$T=1+r(T_z-1)$, where $r=4$. The first latent
step corresponds to the first frame, while each subsequent step
produces four learned temporal phases. Sinusoidal time embeddings and
learned phase embeddings distinguish the resulting frame positions.
The expanded features are refined through spatiotemporal
mixing blocks that combine depthwise spatial convolutions,
per-location temporal self-attention, and channel-wise
feed-forward layers. Finally, an output projection produces
$\widehat{\mathbf{Y}}^{\mathrm{A}}
\in\mathbb{R}^{49\times1225\times1024}$.
\vspace{-15pt}

\paragraph{V-RAE Semantic-Latent Bridge.}
The V-RAE branch operates on a semantically richer latent representation.
A frozen DINOv3 K7 encoder extracts frame-level feature grids, and a frozen temporal pooler compresses each group of four frames into one latent chunk. For a 49-frame video at $560\times560$ resolution, the V-RAE latent has shape $\mathbf{Z}^{\mathrm{V}}\in\mathbb{R}^{13\times1024\times35\times35}$, matching the spatial grid and channel dimension of the VGGT-$\Omega$ tokens.
The target token tensor has shape $\widehat{\mathbf{Y}}^{\mathrm{V}}\in\mathbb{R}^{49\times1225\times1024}$. Thus, temporal expansion is sufficient to match the target dimensions.

To align the conditioning frame with the following 48 frames,
we prepend three copies of the first frame before temporal
pooling, producing a 52-frame input and 13 latent chunks.
For each spatial token, LayerNorm and a linear projection
produce four residuals. Adding each residual and its learned
phase embedding to the original token yields four temporally
ordered tokens, expanding 13 latent chunks into 52 frame features. Because the V-RAE latent already matches the spatial grid size and channel dimension of the VGGT-$\Omega$ tokens, no spatial resampling is required. The bridge applies a residual spatial MLP to align the expanded features with VGGT-$\Omega$ tokens, followed by per-patch temporal attention to model relationships across frames. Finally, the first three padding phases
are removed to recover the 49-frame sequence, yielding
\(\widehat{\mathbf{Y}}^{\mathrm{V}}
\in \mathbb{R}^{49\times1225\times1024}\).

\subsection{Shared Token Alignment Objective}

Both bridges are trained using the same frozen Omega patch tokens as
distillation targets. Only the selected bridge is optimized. All pretrained components remain frozen, and the downstream aggregator and geometry heads are not executed during bridge training.
The alignment objective combines token reconstruction, directional feature
alignment, adjacent-frame dynamics, and sequence-level temporal relations:
\vspace{-5pt}

\begin{equation}
\begin{aligned}
\mathcal{L}_{\mathrm{align}}
={}&
\lambda_{\mathrm{mse}}
\operatorname{MSE}
\left(
\widehat{\mathbf{Y}},
\mathbf{Y}
\right)
+
\lambda_{\mathrm{cos}}
\left[
1-
\operatorname{Cos}
\left(
\widehat{\mathbf{Y}},
\mathbf{Y}
\right)
\right]\\
&+
\lambda_{\Delta}
\operatorname{Huber}
\left(
\Delta_t\widehat{\mathbf{Y}},
\Delta_t\mathbf{Y}
\right)
+
\lambda_{\mathrm{rel}}
\operatorname{MSE}
\left(
\mathcal{R}\left(\widehat{\mathbf{Y}}\right),
\mathcal{R}\left(\mathbf{Y}\right)
\right),
\end{aligned}
\label{eq:shared_alignment_loss}
\end{equation}

where \(\operatorname{MSE}(\cdot,\cdot)\) is the mean squared error,
\(\operatorname{Cos}(\cdot,\cdot)\) is the mean token-wise cosine similarity,
and \(\operatorname{Huber}(\cdot,\cdot)\) denotes the smooth
\(\ell_1\) loss. The temporal difference is defined as $\Delta_t \mathbf{Y}_t = \mathbf{Y}_{t+1} - \mathbf{Y}_t$.

For a token sequence $\mathbf{Y}\in\mathbb{R}^{T\times N\times C}$, we average the tokens within each frame and $\ell_2$-normalize the resulting vectors. Stacking these vectors as rows gives $\mathbf{F}\in\mathbb{R}^{T\times C}$, from which we compute $\mathcal{R}(\mathbf{Y})=\mathbf{F}\mathbf{F}^{\top}\in\mathbb{R}^{T\times T}$. Each entry $R(Y)_{ij}$ measures the cosine similarity between frames $i$ and $j$. The MSE and cosine terms align feature values and directions, respectively. The temporal difference term preserves local feature changes between adjacent frames, while the relation term captures broader temporal relationships across the sequence. Together, these terms encourage feature fidelity and temporal consistency with the teacher. We set $\lambda_{\mathrm{mse}}=\lambda_{\mathrm{cos}}=1$, $\lambda_{\Delta}=0.5$, and $\lambda_{\mathrm{rel}}=0.1$.

\subsection{Direct Geometry Decoding}

At inference, the trained bridge takes the final denoised latent from a compatible video generator as input. Let \(G_e\) denote the generator
associated with representation \(e\). Given a generation condition
\(\mathbf{c}\), such as a reference image and a target camera trajectory, and
a noise variable \(\boldsymbol{\epsilon}\), the generator produces a latent
\(\mathbf{Z}^{g,e}\):
\vspace{-5pt}

\begin{equation}
\begin{aligned}
\mathbf{Z}^{g,e}
    &=G_e(\mathbf{c},\boldsymbol{\epsilon}),\\
\widehat{\mathbf{X}}^{e}
    &=D_e\left(\mathbf{Z}^{g,e}\right),\\
\left(
\widehat{\mathbf{d}}^{e},
\widehat{\mathbf{K}}^{e},
\widehat{\mathbf{E}}^{e}
\right)
    &=
H_{\Omega}
\left(
A_{\Omega}
\left(
B_{\theta_e}
\left(
\mathbf{Z}^{g,e}
\right)
\right)
\right).
\end{aligned}
\vspace{-10pt}
\label{eq:direct_geometry_decoding}
\end{equation}

Here, \(D_e\) denotes the optional RGB decoder of the corresponding video
representation, and \(\widehat{\mathbf{X}}^{e}\) is the generated RGB video.
The frozen Omega aggregator and geometry heads are denoted by \(A_{\Omega}\)
and \(H_{\Omega}\), respectively. Their outputs
\(\widehat{\mathbf{d}}^{e}\),
\(\widehat{\mathbf{K}}^{e}\), and
\(\widehat{\mathbf{E}}^{e}\) represent the predicted depth maps, camera
intrinsics, and camera extrinsics.

We obtain world-space point clouds
$\hat{P}^{e}=\Pi^{-1}(\hat{d}^{e},\hat{K}^{e},\hat{E}^{e})$,
where $\Pi^{-1}$ denotes depth unprojection into a shared world
coordinate system using the predicted camera parameters.
The geometry branch operates directly on generated latents,
bypassing RGB decoding and image feature extraction.

\section{Experiments}
\label{gen_inst}

\paragraph{Data}
We train the camera-conditioned video generation model and the alignment model on a mixture of DL3DV~\citep{ling2024dl3dv}, WildRGB-D~\citep{xia2024rgbd}, CO3Dv2~\citep{reizenstein2021common}, and RealEstate10K~\citep{zhou2018stereo}. To balance their contributions, we sample datasets at a ratio of 2:1:1:1, then randomly sample a contiguous 49-frame video window from an eligible segment in the selected dataset. 
All video frames are resized and cropped to \(560\times560\), with corresponding adjustments to the camera intrinsics. 

\paragraph{Baseline and Metrics.} 
We selected methods that perform simultaneous generation and reconstruction for comparison, including Aether~\citep{zhu2025aether}, FantasyWorld~\citep{dai2026fantasyworld} and Gen3R~\citep{huang2026gen3r}. Aether's maximum generation supports only 41 frames, so its metrics are calculated based on 41 frames. We evaluate the quality of the reconstruction using Accuracy, Completeness, and Chamfer Distance (CD)~\citep{fan2017point}. We evaluate the generated results against the ground truth using PSNR, SSIM~\citep{wang2004image}, and LPIPS~\citep{zhang2018unreasonable}, and assess the model's generation capabilities using the I2V Subject (I2V Subj.), I2V Background (I2V BG), and Imaging Quality (I.Q.) metrics from VBench~\citep{huang2024vbench}.

\paragraph{Implementation Details.} The Wan2.1 bridge is trained for 94K steps with global batch size 32 on 32 GPUs ( NVIDIA RTX PRO 6000, then RTX 5090). The V-RAE bridge is trained for 50K steps using 16 RTX PRO 6000 GPUs. Both bridges use AdamW, BF16, weight decay of \(10^{-4}\), and gradient clipping at 1. Learning rates are \(5\times10^{-5}\) with 1K-step warmup for Wan2.1 and a constant \(5\times10^{-5}\) for V-RAE. These stages cost approximately 1,501 and 498 GPU-hours, respectively.

\begin{figure*}[th]
	\centering
	\includegraphics[width=1\textwidth]{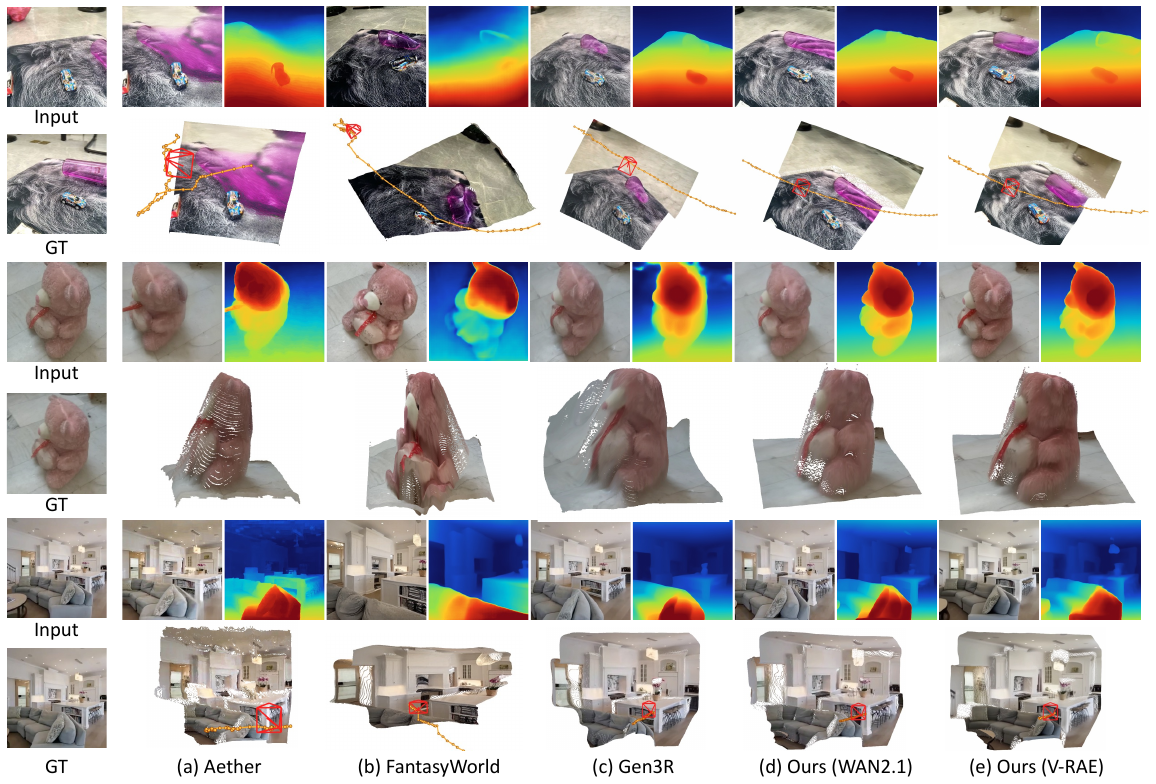}
    \vspace{-20pt}
	\caption{Visual comparison of generation and reconstruction results, including the generated videos, depth maps, and point clouds produced by each method.}
	\label{fig:vis_main}
\end{figure*}

\paragraph{Comparison of Reconstruction Results.} 
We selected 300 Co3Dv2 and 158 WildRGB-D sequences to evaluate the reconstruction performance of different methods. These datasets include ground-truth point clouds. We also present the end-to-end latency from input image to generated video and point cloud. Our comparison method is shown in Tab.~\ref{tab:geometry_comparison}. Our V-RAE variant achieves the lowest Chamfer Distance on both datasets, while the Wan2.1 variant achieves the lowest accuracy error on WildRGB-D. Our Wan2.1 variant records the shortest end-to-end latency
under the evaluated settings. However, differences in
generators and output lengths prevent attributing this
advantage solely to the Bridge. Table~\ref{tab:geometry_generated_real} isolates its
contribution using the same generated latents and frozen
geometry model. Figure~\ref{fig:vis_main} presents our I2V results alongside the corresponding depth maps and 3D visualizations, demonstrating that Latent Bridge recovers sharp depth maps and produces accurate 3D point clouds.

\begin{table*}[t]
\centering
\caption{
Geometry reconstruction results on Co3Dv2 and WildRGB-D.
Geometry metrics are measured in meters and latency in seconds.
The best and second-best results are highlighted in pink and yellow,
respectively. Aether uses 41 frames, whereas the other methods use 49.
}
\label{tab:geometry_comparison}
\resizebox{\textwidth}{!}{
\begin{tabular}{@{}l ccc ccc c@{}}
\toprule
\multirow{2}{*}{Method}
& \multicolumn{3}{c}{Co3Dv2}
& \multicolumn{3}{c}{WildRGB-D}
& \multirow{2}{*}{Latency $\downarrow$} \\
\cmidrule(lr){2-4}
\cmidrule(lr){5-7}
& Accuracy $\downarrow$
& Completeness $\downarrow$
& CD $\downarrow$
& Accuracy $\downarrow$
& Completeness $\downarrow$
& CD $\downarrow$
& \\
\midrule

Aether~\citep{zhu2025aether}
& 0.6982
& 0.5437
& 0.6209
& 0.0591
& 0.1340
& 0.0965
& 181.33 \\

FantasyWorld~\citep{dai2026fantasyworld}
& 0.5939
& 0.3698
& 0.4818
& 0.0440
& 0.1258
& 0.0849
& 448.63 \\

Gen3R~\citep{huang2026gen3r}
& 0.3820
& \cellcolor{pink}\textbf{0.1796}
& 0.2808
& 0.0247
& \cellcolor{yellow}0.1129
& \cellcolor{yellow}0.0688
& 189.57 \\

Ours (Wan2.1)
& \cellcolor{yellow}0.3536
& \cellcolor{yellow}0.1833
& \cellcolor{yellow}0.2685
& \cellcolor{pink}\textbf{0.0189}
& 0.1209
& 0.0699
& \cellcolor{pink}\textbf{70.09} \\

Ours (V-RAE)
& \cellcolor{pink}\textbf{0.3231}
& 0.2093
& \cellcolor{pink}\textbf{0.2662}
& \cellcolor{yellow}0.0237
& \cellcolor{pink}\textbf{0.1076}
& \cellcolor{pink}\textbf{0.0657}
& \cellcolor{yellow}165.57 \\

\bottomrule
\end{tabular}
}
\end{table*}

\vspace{-5pt}
\paragraph{Comparison of generation results.} 
We randomly selected 300 Co3Dv2, 158 WildRGB-D, 300 RealEstate10K, and 140 DL3DV sequences to compare the generation quality. 
Our comparison method is shown in the Tab.~\ref{tab:video_quality_comparison}. In terms of video generation, the Wan2.1 base model produces the highest quality output. V-RAE exhibits slightly lower generation quality than the Wan2.1 VAE. Notably, V-RAE uses a compression ratio of $16\times16\times4$, which is higher than the $8\times8\times4$ compression ratio of the Wan2.1 VAE.
Despite this, V-RAE still outperforms Wan2.1 on the 3D metrics evaluated on CO3D, as its native DINOv3 features facilitate more effective alignment.


\begin{table*}[t]
\centering
\caption{
Video Generation Quality Comparison.
All results are evaluated at $560\times560$ resolution.
I2V Subj., I2V BG, and I.Q. are multiplied by 100.
}
\label{tab:video_quality_comparison}
\setlength{\tabcolsep}{4pt}
\resizebox{\textwidth}{!}{
\begin{tabular}{@{}l cccccc cccccc@{}}
\toprule
\multirow{2}{*}{Method}
& \multicolumn{6}{c}{Co3Dv2}
& \multicolumn{6}{c}{WildRGB-D} \\
\cmidrule(lr){2-7}
\cmidrule(lr){8-13}
& PSNR $\uparrow$
& SSIM $\uparrow$
& LPIPS $\downarrow$
& I2V Subj. $\uparrow$
& I2V BG $\uparrow$
& I.Q. $\uparrow$
& PSNR $\uparrow$
& SSIM $\uparrow$
& LPIPS $\downarrow$
& I2V Subj. $\uparrow$
& I2V BG $\uparrow$
& I.Q. $\uparrow$ \\
\midrule

Aether
& 13.9595 & 0.5689 & 0.5942 & 90.64 & 92.13 & 38.46
& 12.4790 & 0.4466 & 0.5660 & 95.82 & 95.94 & 52.60 \\

FantasyWorld
& 12.0307
& 0.4807
& 0.6741
& 89.08
& 91.27
& 46.20
& 12.9733
& 0.4385
& 0.5116
& 97.07
& 96.72
& \cellcolor{yellow}59.41 \\

Gen3R
& \cellcolor{yellow}16.5289
& \cellcolor{yellow}0.6027
& 0.4405
& 94.79
& \cellcolor{yellow}96.05
& 45.20
& 15.2252
& 0.5075
& 0.4025
& \cellcolor{yellow}98.38
& \cellcolor{yellow}98.20
& 56.58 \\

Ours (V-RAE)
& 16.3780
& 0.6025
& \cellcolor{yellow}0.4303
& \cellcolor{yellow}94.91
& 95.99
& \cellcolor{pink}\textbf{47.84}
& \cellcolor{pink}\textbf{16.5769}
& \cellcolor{pink}\textbf{0.5516}
& \cellcolor{pink}\textbf{0.3217}
& 97.87
& 97.60
& 56.85 \\

Ours (Wan2.1)
& \cellcolor{pink}\textbf{17.0388}
& \cellcolor{pink}\textbf{0.6166}
& \cellcolor{pink}\textbf{0.4121}
& \cellcolor{pink}\textbf{95.42}
& \cellcolor{pink}\textbf{96.41}
& \cellcolor{yellow}46.73
& \cellcolor{yellow}16.1703
& \cellcolor{yellow}0.5412
& \cellcolor{yellow}0.3394
& \cellcolor{pink}\textbf{98.44}
& \cellcolor{pink}\textbf{98.24}
& \cellcolor{pink}\textbf{59.67} \\

\midrule
\multirow{2}{*}{Method}
& \multicolumn{6}{c}{DL3DV}
& \multicolumn{6}{c}{RealEstate10K} \\
\cmidrule(lr){2-7}
\cmidrule(lr){8-13}
& PSNR $\uparrow$
& SSIM $\uparrow$
& LPIPS $\downarrow$
& I2V Subj. $\uparrow$
& I2V BG $\uparrow$
& I.Q. $\uparrow$
& PSNR $\uparrow$
& SSIM $\uparrow$
& LPIPS $\downarrow$
& I2V Subj. $\uparrow$
& I2V BG $\uparrow$
& I.Q. $\uparrow$ \\
\midrule

Aether
& 12.9066 & 0.4211 & 0.5576 & 93.93 & 94.90 & 46.21
& 16.5257 & 0.5865 & 0.3439 & 98.82 & 98.60 & 47.21 \\

FantasyWorld
& 12.1378
& 0.3633
& 0.5944
& \cellcolor{yellow}95.09
& \cellcolor{yellow}95.85
& 47.60
& 14.4308
& 0.5353
& 0.4776
& 97.89
& 97.84
& \cellcolor{pink}\textbf{50.98} \\

Gen3R
& 14.5473
& 0.4234
& 0.4477
& 94.09
& 95.28
& 47.32
& \cellcolor{yellow}19.9467
& \cellcolor{yellow}0.6781
& \cellcolor{pink}\textbf{0.2115}
& \cellcolor{pink}\textbf{99.31}
& \cellcolor{yellow}99.08
& \cellcolor{yellow}49.82 \\

Ours (V-RAE)
& \cellcolor{yellow}14.7968
& \cellcolor{yellow}0.4448
& \cellcolor{yellow}0.4351
& 94.75
& 95.62
& \cellcolor{yellow}48.25
& 17.5092
& 0.6384
& 0.3028
& 98.42
& 97.79
& 47.61 \\

Ours (Wan2.1)
& \cellcolor{pink}\textbf{14.8615}
& \cellcolor{pink}\textbf{0.4453}
& \cellcolor{pink}\textbf{0.4169}
& \cellcolor{pink}\textbf{95.22}
& \cellcolor{pink}\textbf{95.90}
& \cellcolor{pink}\textbf{50.96}
& \cellcolor{pink}\textbf{20.0350}
& \cellcolor{pink}\textbf{0.6917}
& \cellcolor{yellow}0.2125
& \cellcolor{yellow}99.25
& \cellcolor{pink}\textbf{99.22}
& 49.02 \\

\bottomrule
\end{tabular}
}
\end{table*}

\paragraph{Comparison of RGB decoding solutions.} 
We further analyzed the differences between our alignment latent scheme and the RGB decoding and reconstruction scheme, and the results are shown in Tab.~\ref{tab:geometry_generated_real}. For the same generated latents, we compare feed-forward reconstruction via RGB decoding and image encoding with reconstruction through the Bridge, using identical frozen VGGT-$\Omega$ aggregators and prediction heads. The results show that the reconstruction accuracy of the Bridge-based approach is comparable to that of the RGB-decoding pipeline, indicating that the Bridge achieves sufficient fitting. The Bridge preserves comparable reconstruction quality while reducing the measured latent-to-geometry latency from 7.92 s to 4.92 s for Wan2.1 and from 5.76 s to 4.64 s for V-RAE, corresponding to speedups of 1.61× and 1.24×, respectively. 

\vspace{-10pt}

\begin{table*}[h]
\centering
\caption{
Comparison of RGB Decoding Solutions. Latency covers latent-to-geometry inference and excludes diffusion sampling. RGB decoding is included in the RGB
baseline but bypassed by the Bridge.
}
\label{tab:geometry_generated_real}
\resizebox{\textwidth}{!}{
\begin{tabular}{@{}l ccc ccc c@{}}
\toprule
\multirow{2}{*}{Method}
& \multicolumn{3}{c}{CO3Dv2}
& \multicolumn{3}{c}{WildRGB-D}
& \multirow{2}{*}{Time (s) $\downarrow$} \\
\cmidrule(lr){2-4}\cmidrule(lr){5-7}
& Accuracy $\downarrow$ & Completeness $\downarrow$ & CD $\downarrow$
& Accuracy $\downarrow$ & Completeness $\downarrow$ & CD $\downarrow$ & \\
\midrule
Latent $\rightarrow$ RGB $\rightarrow$ VGGT-$\Omega$ (Wan2.1)
& 0.3559 & \textbf{0.1754} & \textbf{0.2657}
& \textbf{0.0174} & 0.1227 & 0.0700 & 7.92 \\
Latent $\rightarrow$ Bridge $\rightarrow$ VGGT-$\Omega$ (Wan2.1)
& \textbf{0.3536} & 0.1833 & 0.2685
& 0.0189 & \textbf{0.1209} & \textbf{0.0699}
& \textbf{4.92} \\
\cmidrule(lr){1-8}
Latent $\rightarrow$ RGB $\rightarrow$ VGGT-$\Omega$ (VRAE)
& 0.3241 & \textbf{0.1925} & \textbf{0.2583}
& \textbf{0.0209} & 0.1114 & 0.0661 & 5.76 \\
Latent $\rightarrow$ Bridge $\rightarrow$ VGGT-$\Omega$ (VRAE)
& \textbf{0.3231} & 0.2093 & 0.2662
& 0.0237 & \textbf{0.1076} & \textbf{0.0657} & \textbf{4.64} \\
\bottomrule
\end{tabular}
}
\end{table*}

\vspace{-10pt}
\section{Ablation study and Generalization Evaluation}

\begin{figure*}[h!]
	\centering
	\includegraphics[width=1\textwidth]{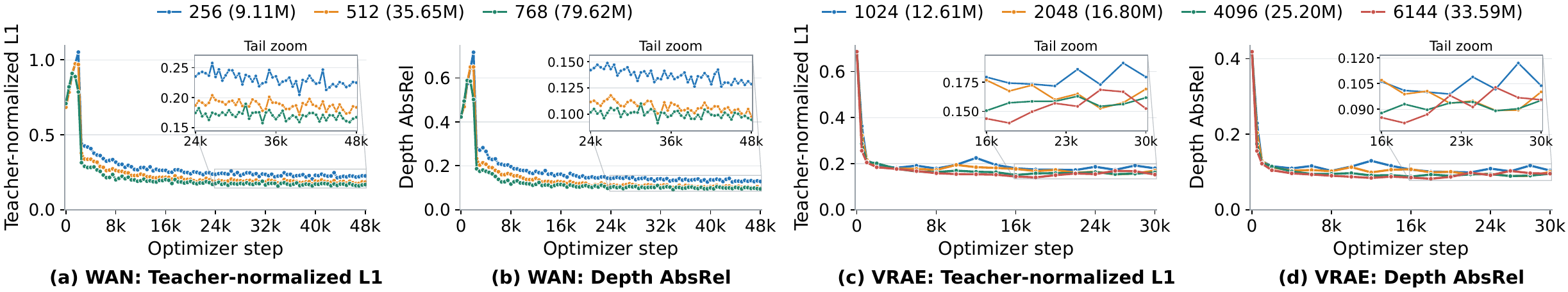}
    \vspace{-15pt}
	\caption{Ablation study on the number of parameters.}
	\label{fig:ablation_parameter}
\end{figure*}

\paragraph{Parameter settings.}
In this ablation study, we investigate how bridge capacity affects latent-space alignment. As shown in Fig.~\ref{fig:ablation_parameter}, we vary only the internal hidden width of the bridge network to control its capacity, keeping all other settings fixed within each model family. For the Wan Bridge, widths of 256, 512, and 768 yield 9.11M, 35.65M, and 79.62M trainable parameters, respectively. For the VRAE Bridge, we vary the hidden width of the spatial mapping MLP and temporal FFN across 1024, 2048, 4096, and 6144, yielding 12.61M, 16.80M, 25.20M, and 33.59M parameters. All alignment variants are trained from scratch on a mixture of four datasets and evaluated for teacher-depth consistency on a fixed set of 64 validation videos. We compute teacher-normalized L1 and AbsRel over all pixels in all 49 frames, without foreground masking or prediction-scale fitting. 

\vspace{-5pt}

\paragraph{Network structure.}
We conduct structural ablations of the Bridge on the Wan branch. Starting from the full Bridge, we separately replace spatial packing with average pooling, causal phase expansion with fixed linear interpolation, and spatiotemporal mixing with a token-wise MLP to assess each component’s contribution. 
All variants are trained from scratch using the same training data, frozen Wan encoder, VGGT-$\Omega$ downstream network, distillation losses, and optimization settings. We additionally introduce a simple baseline in which all three aforementioned components are replaced. We evaluate them at matched training steps on a fixed validation set of 64 videos, each containing 49 frames, using full-image teacher-normalized L1 and depth AbsRel to measure depth distillation consistency. 
As shown in Tab.~\ref{tab:bridge_structure_ablation}, the full Bridge achieves the lowest Depth L1 and AbsRel. Replacing spatial packing with average
pooling slightly improves Norm.\ L1 but worsens the other
two metrics. Replacing causal phase expansion or
spatiotemporal mixing increases all three errors.
Overall, these results support the combined design,
particularly the importance of spatiotemporal mixing
for accurate depth distillation.

\paragraph{Loss function.} We conduct an ablation study on the training loss function. As shown in Tab.~\ref{tab:bridge_structure_ablation}, we refer to the second, third, and fourth terms in ~\eqref{eq:shared_alignment_loss} as the cosine loss, delta loss, and relation loss, respectively. We ablate each loss term and evaluate its fitting capability. As shown in the table, removing any loss term leads to a degradation in accuracy, demonstrating that each constraint contributes to the overall performance. In practice, the weights of the individual loss terms can be adjusted according to the specific characteristics of each Bridge.
\vspace{-10pt}

\begin{table*}[h]
\centering
\caption{
Ablations of the Wan Bridge at 50K optimization steps.
All variants are evaluated on the same 64 clips of 49 frames.
}
\label{tab:bridge_structure_ablation}
\resizebox{\textwidth}{!}{
\begin{tabular}{@{}l ccc  l ccc@{}}
\toprule
Structural ablation
& Norm. L1 $\downarrow$
& Depth L1 $\downarrow$
& AbsRel $\downarrow$
& Loss ablation
& Norm. L1 $\downarrow$
& Depth L1 $\downarrow$
& AbsRel $\downarrow$ \\
\midrule
w/o $2\times2$ PixelUnshuffle
& \cellcolor{pink}{\textbf{0.18057}} & \cellcolor{yellow}{0.15849} & 0.10628
& w/o Cosine loss
& 0.18823 & 0.16459 & 0.10851 \\

w/o Causal Conv3D + phase rearrangement
& 0.19177 & 0.16593 & 0.10766
& w/o Delta loss
& 0.18780 & 0.16235 & 0.10461 \\

w/o Spatiotemporal mixing blocks
& 0.25196 & 0.22100 & 0.15228
& w/o Relation loss
& 0.18412 & 0.15863 & \cellcolor{yellow}{0.10313} \\
\midrule
Simple baseline (joint replacements)
& 0.31666 & 0.27490 & 0.19346&
\textbf{Full Bridge \& loss}
& \cellcolor{yellow}{0.18133} & \cellcolor{pink}{\textbf{0.15644}} & \cellcolor{pink}{\textbf{0.10163}} \\
\bottomrule
\end{tabular}
}
\end{table*}
\vspace{-10pt}

\begin{figure*}[th]
	\centering
	\includegraphics[width=1\textwidth]{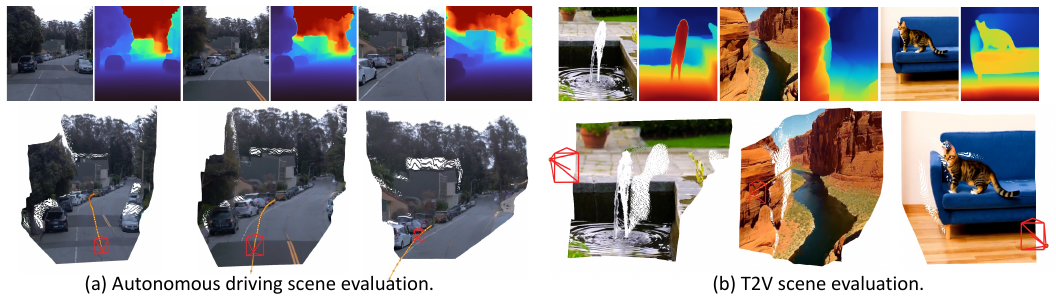}
    \vspace{-20pt}
	\caption{The Bridge can adapt to generative models for different tasks that use the same VAE, without retraining.}
	\label{fig:vis}
    \vspace{-10pt}
\end{figure*}

\paragraph{Scene generalization evaluation.} The alignment module can be transferred across different generative models trained using the same VAE. As shown in Fig.~\ref{fig:vis}(a), we utilized the Waymo dataset~\citep{sun2020scalability} to train a generative model for driving scenarios, based on the Wan 2.1 VAE. The alignment network can be directly used to decode 3D output without the need for retraining.
\vspace{-10pt}

\paragraph{T2V model generalization evaluation.} To verify the reusability of the geometry interface, we directly integrate the existing Bridge with the official Wan2.1-T2V-1.3B model. The generator operates in the latent space of its accompanying Wan2.1 VAE, allowing its final denoised latents to be fed directly into Latent Bridge without RGB re-encoding. The frozen VGGT-$\Omega$ then predicts depth and camera parameters. This integration requires no additional T2V-specific training, parameter updates, or test-time optimization. As shown in Fig.~\ref{fig:vis}(b), the experiment demonstrates that the geometry decoding interface can be reused with another generator sharing a compatible VAE latent space, without retraining the generative model.

\section{Conclusion}

We presented Latent Bridge, a framework that connects video generation and feed-forward 3D reconstruction through latent-space feature alignment. By mapping Wan2.1 and V-RAE video representations to a predefined geometric feature interface, our method enables direct prediction of cameras, depth, and point clouds while keeping the downstream geometric model frozen. This design bypasses RGB decoding and visual re-encoding in the geometry branch, allowing a shared video latent representation to support both video synthesis and geometry prediction. Experiments on Co3Dv2 and WildRGB-D demonstrate competitive reconstruction quality with improved inference efficiency. More broadly, our results suggest that feature alignment offers a practical route to reusing pretrained geometric models with generative video representations.

\subsection*{AI use statement}

In this work, we used generative AI tools to assist with writing and language polishing, including refining English expressions. We also used these tools for literature retrieval and discovery to identify potentially overlooked references. In addition, we used them to support research ideation and execution, particularly in writing visualization scripts and developing a 3D visualization interface. In addition, we used AI to review drafts, identify potential issues, and provide suggestions.
We have not used generative AI tools to directly write any sections of the paper, and generating synthetic datasets and proving mathematical claims are not applicable to this work.
We have reviewed all AI-assisted work. We verified the accuracy of the related-work information provided by generative AI and reviewed and validated the generated code. We take responsibility for the final content of this work, including text, claims or artifacts produced with the aid of generative AI.


\bibliography{iclr2027_conference}
\bibliographystyle{iclr2027_conference}

\clearpage
\appendix
\section{Appendix}

In the appendix, we provide further methodological and experimental details. In Section~\ref{supp.a1}, we provide a detailed comparison between our method and existing approaches. In Section~\ref{supp.a2}, we examine downstream reuse for cross-view segmentation by connecting a VGGT-aligned Bridge to a frozen pretrained segmentation module. In Section~\ref{supp.a3}, we provide detailed information on data usage and evaluation protocols. In Section~\ref{supp.a4}, we discuss the limitations of our method. 

\subsection{Comparison with Existing Methods}
\label{supp.a1}
As shown in Tab.~\ref{tab:method_comparison}, we compare our method with existing approaches from two perspectives: compatibility with existing models and training overhead. Cross-DiT checkpoint reuse refers to whether the method can produce 3D outputs using different generative DiT checkpoints without retraining. FantasyWorld trains a geometry branch and cross-branch adapters while keeping the video backbone frozen, whereas Gen3R fine-tunes the video DiT to jointly generate appearance and geometry latents. Beyond Pixels and our method instead learn geometry pathways from video latents without updating the DiT, enabling reuse across compatible generators that share the same VAE checkpoint and latent conventions. We demonstrate this reuse by applying the same Wan Bridge checkpoint to camera-conditioned generation and the official Wan2.1-T2V-1.3B model without additional Bridge training. Pretrained task-head reuse refers to the ability to reuse existing downstream heads because our method outputs features in VGGT’s DINO feature space. Consequently, the resulting features can be used not only for reconstruction with VGGT but also for cross-view segmentation with VGGT-S, as shown in Section~\ref{supp.a2}. Gen3R reconstructs intermediate VGGT geometry tokens, which are decoded by pretrained geometry heads. In contrast, FantasyWorld and Beyond Pixels incorporate their decoder heads into training and therefore do not offer this advantage. Therefore, our method offers substantially greater versatility than existing approaches.

In addition, we compare the additional training costs required to support both generation and reconstruction. FantasyWorld trains a geometry branch and cross-branch adapters, while Gen3R fine-tunes the video DiT for joint video and geometry generation. Beyond Pixels trains an alignment module, LoRA adapters, and geometry prediction heads. In contrast, we train only the Bridge through feature alignment, keeping the video encoder, geometric aggregator, and prediction heads frozen. Keeping the decoder heads frozen further reduces the training cost. Furthermore, FantasyWorld and Gen3R introduce additional geometry tokens. Our bridge is trained using feature supervision alone, without executing the downstream geometric aggregator or prediction heads during optimization.

\begin{table*}[h]
\centering
\caption{Differences Between Our Method and Existing Approaches.
}
\label{tab:method_comparison}
\resizebox{\textwidth}{!}{
\begin{tabular}{@{}l ccc ccc@{}}
\toprule
\multirow{2}{*}{Method}
& \multicolumn{3}{c}{Pretrained Checkpoint Reuse without Further Training}
& \multicolumn{2}{c}{Trainable Modules and Token Overhead.}
 \\
\cmidrule(lr){2-4}\cmidrule(lr){5-6}
& Cross-DiT checkpoint & T2V/I2V checkpoint & Pretrained task-head
& Trainable components & Additional Geometry Tokens  \\
\midrule

FantasyWorld
& No
& No
& No
& Geometry branch \& adapters
& Yes
 \\

Gen3R
& No
& No
& Yes
& DiT \& geometry adapter
& Yes
 \\

Beyond Pixels
& Yes
& Yes
& No
& Alignment module, LoRA \& heads
& No
 \\

VIST3A
& Yes
& -
& No
& Stitching layer \& geometry/DiT LoRA
& No\\

Latent Bridge
& Yes
& Yes
& Yes
& Bridge only
& No\\

\bottomrule
\end{tabular}
}
\end{table*}

\begin{table}[t]
\centering
\caption{
Computational cost of different supervision strategies.
All settings use the same 6000 Pro GPU with a batch size of 1
and 49 frames at $560\times560$ resolution.
}
\label{tab:supervision_cost}
\resizebox{0.85\linewidth}{!}{
\begin{tabular}{@{}llccc@{}}
\toprule
Supervision
& Trainable modules
& Params (M) $\downarrow$
& Time (s) $\downarrow$
& Memory (GiB) $\downarrow$ \\
\midrule
Feature (ours)
& Bridge
& \textbf{35.65}
& \textbf{0.166}
& \textbf{3.17} \\

Geometry
& Bridge
& \textbf{35.65}
& 5.896
& 31.27 \\

Geometry
& Bridge + geometry model
& 876.31
& 6.375
& 35.48 \\
\bottomrule
\end{tabular}
}
\end{table}

To compare the alignment efficiency of feature supervision with that of geometry supervision provided through the frozen decoder heads, we conduct an experiment, as shown in Table~\ref{tab:supervision_cost}.
Our feature-supervised bridge requires only 0.166\,s and 3.17\,GiB per forward--loss--backward pass, compared with 5.896\,s and 31.27\,GiB for geometry supervision with frozen VGGT-$\Omega$. Unfreezing its aggregator and camera/depth heads further increases the cost to 6.375\,s and 35.48\,GiB. This demonstrates that feature distillation before the frozen decoder heads is substantially less computationally expensive than applying supervision after the features have passed through the frozen decoder heads.

\subsection{Downstream Reuse Beyond Geometry.} 
\label{supp.a2}
\begin{figure*}[h]
	\centering
	\includegraphics[width=1\textwidth]{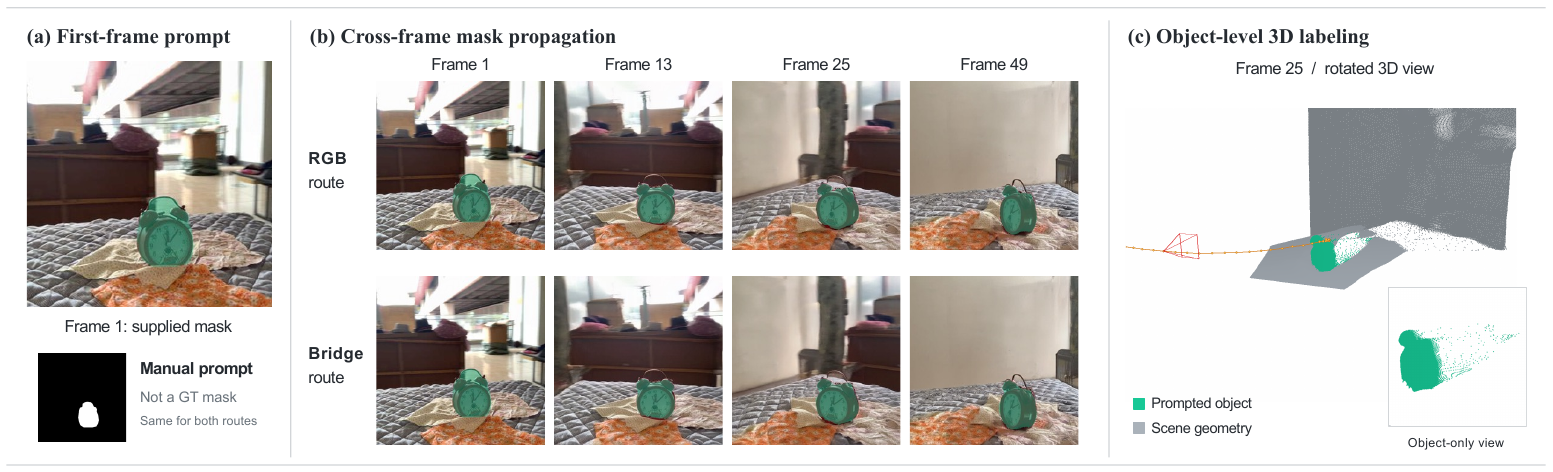}
    \vspace{-20pt}
	\caption{Downstream reuse with a frozen VGGT-S module. Given the same first-frame mask prompt, the RGB and Bridge routes propagate object masks across generated frames. Predicted geometry enables object-level 3D labeling.}
	\label{fig:subb_seg1}
    \vspace{-10pt}
\end{figure*}

\begin{figure*}[h]
	\centering
	\includegraphics[width=0.9\textwidth]{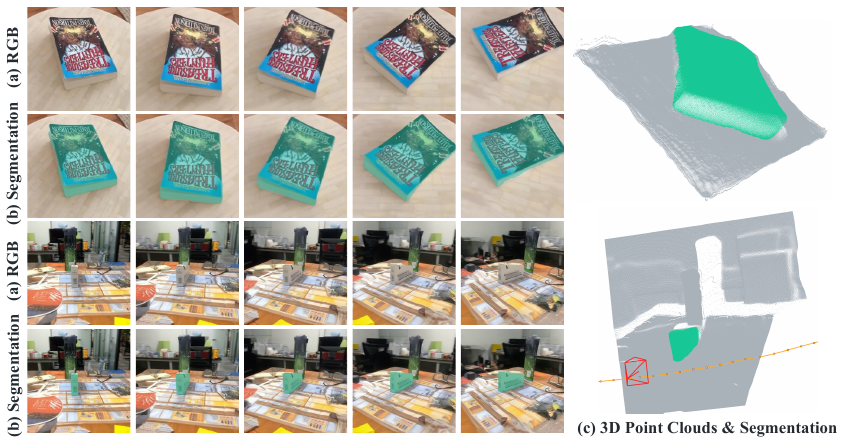}
    \vspace{-10pt}
	\caption{Additional downstream reuse examples showing generated RGB frames, propagated object masks, and the corresponding label-colored 3D point clouds.}
	\label{fig:subb_seg2}
    \vspace{-10pt}
\end{figure*}

Beyond geometry reconstruction, we extend our framework to the standard VGGT feature space and examine downstream reuse. This experiment uses a separately trained VGGT-aligned Bridge, rather than the VGGT-$\Omega$ Bridge evaluated above. Within the VGGT-aligned branch, the Bridge checkpoint used for geometry reconstruction also supports cross-view segmentation through a frozen pretrained VGGT-S~\citep{Gao_2026_CVPR} module. This reuse requires only resolution and feature-interface adjustments, without additional training or fine-tuning. As shown in Fig.~\ref{fig:subb_seg1}, given a first-frame object mask, the pipeline propagates masks across generated frames without decoding the generated video into RGB. We present additional results in Fig.~\ref{fig:subb_seg2}. Associating these masks with the predicted 3D coordinates further produces label-colored object point clouds. These results demonstrate task reuse within the VGGT-aligned branch, beyond its original geometry application.

\subsection{Datasets and Evaluation Protocol.} 
\label{supp.a3}
\paragraph{Datasets.} Table~\ref{tab:dataset_splits} summarizes dataset usage and
split sizes. Counts refer to unique source scenes, sequences,
or video clips rather than sampled training windows.
We follow the official scene lists for DL3DV, the official
novel-view-synthesis split for WildRGB-D, and the official
train/test split for RealEstate10K (RE10K). For CO3Dv2, we
construct a category-stratified 90/10 split at the sequence
level. Evaluation uses 140 DL3DV scenes, 158 valid WildRGB-D
scenes, 300 CO3Dv2 sequences, and 300 RE10K clips.
For WildRGB-D, we exclude two of the 160 evaluation candidates
because they contain no valid global ground-truth points.

\begin{table*}[h]
\centering
\caption{
Dataset usage and splits.
Counts refer to unique source scenes, sequences, or video clips,
rather than sampled training windows.
}
\label{tab:dataset_splits}
\resizebox{\textwidth}{!}{
\begin{tabular}{@{}l l c c c l@{}}
\toprule
Dataset
& Usage
& Training
& Validation Pool
& Test
& Split Basis \\
\midrule
DL3DV
& Training; video evaluation
& \shortstack{6,209}
& 140
& 140
& Official scene lists with validity filtering \\

WildRGB-D
& Training; geometry \& video evaluation
& 5,786
& 160
& 158
& Official NVS train/validation split \\

CO3Dv2
& Training; geometry \& video evaluation
& 19,154
& 2,111
& 300
& Custom category-stratified, sequence-level 90/10 split \\

RE10K
& Training; video evaluation
& 57,240
& 6,331
& 300
& Official train/test split; seed-42 test sampling \\

\bottomrule
\end{tabular}
}
\par\smallskip
\end{table*}

\paragraph{Geometry metrics.}
We evaluate reconstruction using Accuracy, Completeness, and Chamfer Distance (CD), with lower values indicating better results.
Accuracy measures the mean distance from each predicted point to its nearest GT point:
\begin{equation}
\mathrm{Accuracy}(P,G)
= \frac{1}{|P|}\sum_{p\in P}\min_{g\in G}\|p-g\|_2.
\end{equation}
Completeness measures the reverse distance, quantifying how well the prediction covers the GT geometry:
\begin{equation}
\mathrm{Completeness}(P,G)
= \frac{1}{|G|}\sum_{g\in G}\min_{p\in P}\|g-p\|_2.
\end{equation}
We define CD as the arithmetic mean of these two directional errors:
\begin{equation}
\mathrm{CD}(P,G)
= \frac{\mathrm{Accuracy}(P,G)+\mathrm{Completeness}(P,G)}{2}.
\end{equation}
All distances are Euclidean rather than squared.
Metrics are computed on the merged point cloud of each sequence and then averaged with equal weight per sequence within each dataset.

\paragraph{Depth filtering.}

Depth filtering removes potentially unreliable predicted points and is distinct from foreground segmentation.
Candidate pixels are first restricted by the supplied dataset masks and GT-depth validity checks.
For each frame, we compute the median and median absolute deviation of finite positive predicted log depths:
\begin{equation}
m=\operatorname{median}_{i}(\log d_i),
\qquad
a=\operatorname{median}_{i}|\log d_i-m|.
\end{equation}
Log-MAD filtering retains candidates satisfying
\begin{equation}
|\log\max(d_i,10^{-8})-m|
\leq 3\times1.4826\,a.
\end{equation}
If $a\leq10^{-8}$, all finite positive candidates are retained.
We then check the surviving points for multi-view consistency by back-projecting them into 3D and projecting them into available neighboring frames at offsets $\{-2,-1,1,2\}$ using the reconstruction cameras.
A neighboring view supports a point when its rounded projected pixel is in bounds, both depths are valid, and
\begin{equation}
\frac{|d_{\mathrm{obs}}-d_{\mathrm{proj}}|}
{\max(d_{\mathrm{obs}},d_{\mathrm{proj}},10^{-8})}
\leq 0.10.
\end{equation}
Only points supported by at least two neighboring views are retained.
Neighboring depth observations are read directly from the predicted depth maps without requiring them to pass their own foreground or log-MAD masks.
Together, these filters suppress depth outliers and cross-view inconsistencies. All methods use the same depth-filtering procedure and thresholds, ensuring consistency in filtering across methods for a fair comparison.

\subsection{Limitation.} 
\label{supp.a4}
Our Bridge network has the following limitations:
First, our model depends on a pretrained geometry model.
The bridge learns to match teacher features while keeping the downstream geometry model frozen, and therefore cannot be expected to correct the teacher's geometric errors. Second, inconsistencies in generated videos are not automatically corrected. Our method extracts geometry from generated latents but does not explicitly constrain the video generation process. Therefore, if the generated content exhibits structural changes or object inconsistencies across frames, the Bridge cannot guarantee the reconstruction of a coherent 3D scene.

\end{document}